\documentclass[letterpaper, 10 pt, conference]{ieeeconf}  % Comment this line out if you need a4paper

\IEEEoverridecommandlockouts  
\usepackage{amsfonts}
\usepackage{amsmath}
\usepackage[dvipsnames]{xcolor}

\usepackage{algorithm,algorithmic}
\usepackage{graphicx}
\graphicspath{ {./figs/} }
\usepackage{caption}
\usepackage{subcaption}

\usepackage{bbm}
\usepackage{amssymb}

\title{\LARGE \bf
Deformable Linear Object Prediction Using Locally Linear Latent Dynamics
}

\author{Wenbo Zhang$^{1}$, Karl Schmeckpeper$^{2}$, Pratik Chaudhari$^{3}$, and Kostas Daniilidis$^{4}$% <-this % stops a space
\thanks{$^{1}$Wenbo Zhang did the work when he was a student at the GRASP Lab, University of Pennsylvania, USA {\tt\small zwenbo@seas.upenn.edu}}%
\thanks{$^{2}$Karl Schmeckpeper is with the GRASP Lab, University of Pennsylvania, USA {\tt\small karls@seas.upenn.edu}}%
\thanks{$^{3}$Pratik Chaudhari is with the GRASP Lab, University of Pennsylvania, USA {\tt\small pratikac@seas.upenn.edu}}%     
\thanks{$^{4}$Kostas Daniilidis is with the GRASP Lab, University of Pennsylvania, USA {\tt\small kostas@cis.upenn.edu}}%   
\thanks{Code: https://github.com/zwbgood6/deform}% 
}

\begin{document}

\maketitle
\thispagestyle{empty}
\pagestyle{empty}

%%%%%%%%%%%%%%%%%%%%%%%%%%%%%%%%%%%%%%%%%%%%%%%%%%%%%%%%%%%%%%%%%%%%%%%%%%%%%%%%
\begin{abstract}
We propose a framework for deformable linear object prediction. Prediction of deformable objects (e.g., rope) is challenging due to their non-linear dynamics and infinite-dimensional configuration spaces. By mapping the dynamics from a non-linear space to a linear space, we can use the good properties of linear dynamics for easier learning and more efficient prediction. We learn a locally linear, action-conditioned dynamics model that can be used to predict future latent states. Then, we decode the predicted latent state into the predicted state. We also apply a sampling-based optimization algorithm to select the optimal control action. We empirically demonstrate that our approach can predict the rope state accurately up to ten steps into the future and that our algorithm can find the optimal action given an initial state and a goal state. 
\end{abstract}

%%%%%%%%%%%%%%%%%%%%%%%%%%%%%%%%%%%%%%%%%%%%%%%%%%%%%%%%%%%%%%%%%%%%%%%%%%%%%%%%
\section{INTRODUCTION}
Robotic manipulation can be applied to a variety of scenarios, such as surgery, housekeeping, manufacturing, and agriculture~\cite{sanchez2018robotic}\cite{khalil2010dexterous}. Deformable object manipulation is one of the important robotic manipulation problems.
We face deformable objects frequently in our everyday lives, such as when folding the clothes~\cite{doumanoglou2016folding}, knot-typing~\cite{saha2006motion}\cite{wang2016tying}, and moving a cable~\cite{she2019cable}.
However, deformable object manipulation is still a challenging task due to the following reasons. (i) High model uncertainty. 
It is hard to simulate an accurate deformable object model because of environmental noises. (ii) Infinite-dimensional configuration spaces. We need to get lots of parameters for modeling a good deformable object. (iii) High cost of simulation. It is expensive to get all the simulation model parameters for deformable objects, and sometimes it is infeasible to have all the parameters. (iv) Hyper-underactuated systems. The robotic manipulator cannot fully control deformable objects with arbitrary trajectories. (v) Non-linear dynamics. Non-linearity makes it hard to predict future states when executing actions on deformable objects. Our paper explicitly addresses the problems caused by non-linear dynamics and infinite-dimensional configuration spaces through mapping dynamics from a non-linear space to a linear space and shrinking the state to a lower-dimensional latent state.

In this paper, we consider the problems on how to predict the future state of deformable linear objects (i.e., rope) given an initial state and a sequence of actions, and on how to find optimal control actions to move the rope from the initial state to a goal state given a prediction model.

To solve the prediction problem, we propose a prediction framework that consists of a state autoencoder model, an action autoencoder model, and a dynamics model. The state autoencoder model is based on convolutional neural networks.
Rope dynamics are non-linear and are very complicated to learn directly.
Unlike non-linear dynamics, linear dynamics are easier to learn and more efficient for state predictions. 
Thus, we consider mapping the non-linear space into a linear space, predicting future states in the linear space, and decoding back to the non-linear space from the linear space. 
To do so, we first use learned encoders to encode both the state and the action into the latent state and the latent action. Then, we run the dynamics model to get the state matrix and control matrix for the linear dynamics in the latent space. After that, we get a predicted next latent state by using locally linear latent dynamics based on the current latent state, latent action, state matrix, and control matrix. Finally, we decode the predicted next latent state into a predicted next state. Our experiments show that the prediction framework can accurately predict states up to ten timesteps in the future. 

We also want to find the optimal control actions to manipulate the rope from an initial position to a goal position. To solve this control problem, we use a sampling-based optimization algorithm to select the optimal action sequence based on an initial state, a goal state, and the trained prediction framework. This algorithm is an extension of the cross-entropy method. The algorithm works by repeating the processes of (i) sampling action sequences from a probability distribution, (ii) predicting the goal state, (iii) calculating and sorting the loss between predicted state and ground truth, and (iv) fitting a new distribution to selected action sequences. We stop this process when the distribution converges and select the optimal action sequence with the lowest loss.

\textbf{Contributions.} Our key contributions are: (i) a novel prediction framework for predicting future states of deformable linear objects under a sequence of actions (Section \ref{sec:method}), (ii) a sampling-based optimization algorithm for selecting optimal control actions to move the deformable linear objects (Section \ref{sec:method}), (iii) experiments showing multi-step prediction results and comparing our method with a baseline qualitatively and quantitatively (Section \ref{sec:experiment}), and (iv) ablation studies regarding locally and globally latent dynamics, model training methods, and how small latent size we can use to achieve good performance (Section \ref{sec:experiment}).  

\section{RELATED WORK}
\textbf{Learning Models for Non-linear Systems.} There are active studies about learning a model for non-linear dynamical systems. %Ghahramani et al.~\cite{ghahramani1999learning} apply the classical Expectation-Maximization algorithm to solve the nonlinear dynamical system. Otto et al.~\cite{otto2019linearly} apply linearly-recurrent autoencoder to learn dynamics. 
Kingma et al.~\cite{kingma2014stochastic} introduce the variational autoencoder (VAE) using approximate Bayesian inference. Based on VAE, Watter et al.~\cite{watter2015embed} propose an Embed to Control (E2C) method to learn a non-linear dynamical system model. They use the VAE to learn latent parameters and set locally linear constraints on dynamics using an optimal control concept. We are inspired by this method. We do experiments using locally linear latent dynamics and compare it with globally linear latent dynamics. However, instead of applying the optimal control formulation, we train an action-conditioned dynamics model to enable system identification in the latent space. Our dynamics model can generate the state matrix and the control matrix at each time step. Li et al.~\cite{li2019learning} apply the Koopman operator theory to provide nonlinear-to-linear transformation. Based on this approach, the infinite-dimensional state space changes to finite-dimensional state space. We learn from this method to encode high-dimensional state space into a smaller latent state in the linear space. There are also many researchers~\cite{krishnan2017structured}\cite{karl2016deep}\cite{yeung2019learning} working on learning high-dimensional non-linear state space and system identification. %Krishnan et al.~\cite{krishnan2017structured} build a structured inference network to model nonlinear state space. Karl et al.~\cite{karl2016deep} introduce deep variational Bayes filters to learning state space from raw data. Yeung et al.~\cite{yeung2019learning} use the deep neural networks to learn representations for the Koopman operator.  %There are also other related works using neural networks to solve dynamics problems. \cite{lusch2018deep}\cite{gao2016linear}\cite{chung2015recurrent} 

\textbf{State Estimation and Prediction.} For state estimation, Yan et al.~\cite{yan2020self} apply self-supervised learning for state estimation. Schulman et al.~\cite{schulman2013tracking} incorporate the point cloud in the probabilistic generative model for tracking deformable objects. 
%Tang et al.~\cite{tang2016robotic} introduce a tangent space mapping algorithm to maintain structural information when applying non-rigid registration. 
Petit et al.~\cite{petit2015real} track objects with point cloud data from an RGB-D camera. 
%Doumanoglou et al.~\cite{doumanoglou2016folding} extract features from a depth map to select the grasping point. 
Unlike those approaches to getting the object's points for state estimation, we use the binary mask of rope image as an input state to a state autoencoder. Then, the encoder can output a latent state. For the prediction of state, Finn et al.~\cite{finn2016unsupervised}\cite{finn2017deep} present deep action-conditioned video prediction models. Unlike \cite{finn2016unsupervised}\cite{finn2017deep}, we predict the future states in the latent space based on the estimation of the latent states. Besides, we predict the next state using the current state information without including previous states'.

%%KS.10.31 - To that end, it might be helpful to add some stuff about video prediction/how it can be used for control.  ie Chelsea/Sergey's Deep Visual Foresight.  We are also using a very similar control setup, so we definitely need to cite that line of work.

\textbf{Planning and Control.} The work~\cite{yan2020self} uses Model Predictive Path Integral (MPPI) controller for choosing the best grasping point with the least moving length to the goal. Wang et al.~\cite{wang2019learning} apply the data-driven approach to generate a sequence of planned images and then use an inverse dynamics model to execute the plan. Sundaresan et al.~\cite{sundaresan2020learning} use dense depth object descriptors to learn a policy for manipulating a rope into different configuration settings. Pathak et al.~\cite{pathak2018zero} learn a goal-conditional control policy for the rope manipulation based on a sequence of images. Moll et al.~\cite{moll2006path} compute the minimal-energy curves for path planning in a wire manipulation task. Bayazit et al.~\cite{bayazit2002probabilistic} introduce a Probabilistic Roadmap (PRM) method. 
There are also some other sampling-based methods which have been introduced in~\cite{saha2006motion}\cite{finn2017deep}\cite{anshelevich2000deformable}\cite{lamiraux2001planning}\cite{roussel2015manipulation}. Similar to \cite{finn2017deep}, we have a similar control setup and we also use a sampling-based optimization algorithm to select the optimal control action. Besides, researchers apply imitation learning to learn control policies. Nair et al.~\cite{nair2017combining} use human demonstrations to teach the robot how to move a rope. The robot will learn the policy from the video sequence and apply this policy for execution. Huang et al.~\cite{huang2015leveraging} approach the problem by trajectory transfer through non-rigid registration.

\textbf{Deformable Objects Manipulation.} There are mainly two types of deformable objects. One is about linear objects \cite{roussel2015manipulation} (e.g., rope, cable) and the other is about non-linear objects \cite{doumanoglou2016folding}\cite{li2018model} (e.g., fabric, cloth, bed sheet, garment). 
%Doumanoglou et al. \cite{doumanoglou2016folding} build a whole pipeline for folding and unfolding clothes in several specifies steps. Li et al. \cite{li2018model} use model-driven approach to build 3-D database for deformable garment. Roussel et al. \cite{roussel2015manipulation} use gripper to manipulate an extensible elastic rod without any collision. 
This paper is focusing on a simpler scenario on deformable linear objects, such as the ropes. Besides, our attentions are more on how to predict the future states and how to sample actions to reach a goal state, and less on how to manipulate the deformable objects using robots.

\section{METHOD}
\label{sec:method}
Deformable objects modeling \cite{petit2015real}\cite{li2018model}\cite{gibson1997survey}\cite{essahbi2012soft} is a good way to model the objects in a simulation.
%%KS.10.31 - This paragraph is a little confusing.  Are you saying that existing methods allow for simulating deformable objects, but that these methods are not practical?  Do they fail to go fast enough?  Are their results bad?
However, existing methods are not practical for real-time robotic tasks due to slow speed, high simulation expense, high-dimensional state space, high model uncertainty, and non-linear dynamics. These reasons result in a challenging task of predicting deformable objects in future states under specific actions. Thus, we want to simplify the non-linear dynamics and reduce high-dimensional state space for deformable linear objects. 
%%KS.10.31 - Reducing to locally linear solves the non-linear dynamics, but you might want to mention if it helps on the other problems.

Instead of using a physics-based simulation to model the objects, we design an encoding network to extract the latent states from rope images. Then we learn the dynamical system in the latent space. 
Using the dynamics in the latent space, we can predict the future latent states. After that, the model decodes the future latent states into rope images. We elaborate on each step as following.

\subsection{Preliminary}
First, we introduce some fundamental concepts.

\textbf{Rope Dynamics.}
We consider nonlinear rope dynamics
\begin{equation*}
    x(t+1) = f(x(t), u(t)),
\end{equation*}
where $t \in \{0,1,...,T\}$ is the time step, $x(t) \in \mathcal{X} \subseteq \mathbb{R}^{m}$ is the state, $u(t) \in \mathcal{U} \subseteq \mathbb{R}^{n}$ is the control input. Since we cannot get the rope state easily, we denote the rope image as the state $x(t)$. In our case, $u = (x, y, l, \theta)$, where $x$ and $y$ are positions in the image space, moving length $l$ and moving angle $\theta$ are in the world space.

\textbf{Latent Dynamics.}
The rope dynamics are non-linear, very complicated to learn, and hard to generalize, while linear dynamics are generally simpler to learn and can help make state predictions more easily.
Thus, we consider a linear dynamics model
\begin{equation}
    g(t+1) = K(t)g(t) + L(t)a(t)
    \label{eq:latent_dyn}    
\end{equation}
in the latent space. Here, $g(t) \in \mathcal{G} \subseteq \mathbb{R}^{p}$ is the latent state, $a(t) \in \mathcal{A} \subseteq \mathbb{R}^{q}$ is the latent action, $K(t) \in \mathcal{K} \subseteq \mathbb{R}^{p \times p}$ is the state matrix, and $L(t) \in \mathcal{L} \subseteq \mathbb{R}^{p \times q}$ is the control matrix. The state matrix $K(t)$ is a transition matrix from the current latent state $g(t)$ to the next latent state $g(t+1)$. The control matrix $L(t)$ learns the impact of the latent action $a(t)$ on the state transition.%The control matrix $L(t)$ is a coefficient matrix and it considers the impact of the latent action $a(t)$ on the state transition.

For globally linear latent dynamics, $K(t)=K$ and $L(t)=L$ for all $t \in \{0,1,...,T\}$, where $K$ and $L$ are fixed matrices. For locally linear latent dynamics, $K(t)$ depends on the state $x(t)$ and $L(t)$ depends on both state $x(t)$ and action $u(t)$.

\textbf{Encoder Models.} 
The state encoder model is a map $\phi_e:\mathcal{X}\to\mathcal{G}$, the action encoder model is a map $\varphi_e:\mathcal{U}\to\mathcal{A}$. We then have
\begin{equation*}
    g(t) = \phi_e (x(t)),~a(t) = \varphi_e (u(t)). 
\end{equation*}
Encoder models encode the state $x$ into the latent state $g$ and the action $u$ into the latent action $a$. For the state $x$ (i.e., raw image), it is high-dimensional and redundant, so we want to reduce the state size and use the smaller latent state as a representation. For the action $u$ (i.e., grasping positions, gripper moving length, and gripper moving angle), the raw action usually has a very non-linear effect on the state, so we lift the actions to a higher dimension and make them easier to linearize in the latent space.
%%KS.10.31 - Why do we want to change the size of the representations? For the image, the raw image is very high dimensional, but also very redundant, so we can reduce the size.  For the action, the raw action has very non-linear effects on the state, so by lifting to a higher dimension, we can make them easier to linearize - done

\textbf{Decoder Model.}
The state decoder model is a map $\phi_d:\mathcal{G}\to\mathcal{X}$, the action decoder model is a map $\varphi_d:\mathcal{A}\to\mathcal{U}$. We then have
\begin{equation*}
    \hat{x}(t) = \phi_d (g(t)),~\hat{u}(t) = \varphi_d (a(t)).
\end{equation*}
Paired with encoder models, decoder models decode the latent state $g$ into the reconstructed state $\hat{x}$ and the latent action $a$ into the reconstructed action $\hat{u}$.

\textbf{Dynamics Model.}
The dynamics model contains one map $\psi_s:\mathcal{X}\to\mathcal{K}$ for the state matrix and the other map $\psi_c:\mathcal{X},\mathcal{U}\to\mathcal{L}$ for the control matrix. We then have
\begin{equation*}
    K(t) = \psi_s ( x(t) ),~L(t) = \psi_c ( x(t), u(t) ). 
\end{equation*}

\subsection{Prediction Framework}
\begin{figure}[!ht]
    \centering
    \includegraphics[width=0.65\columnwidth]{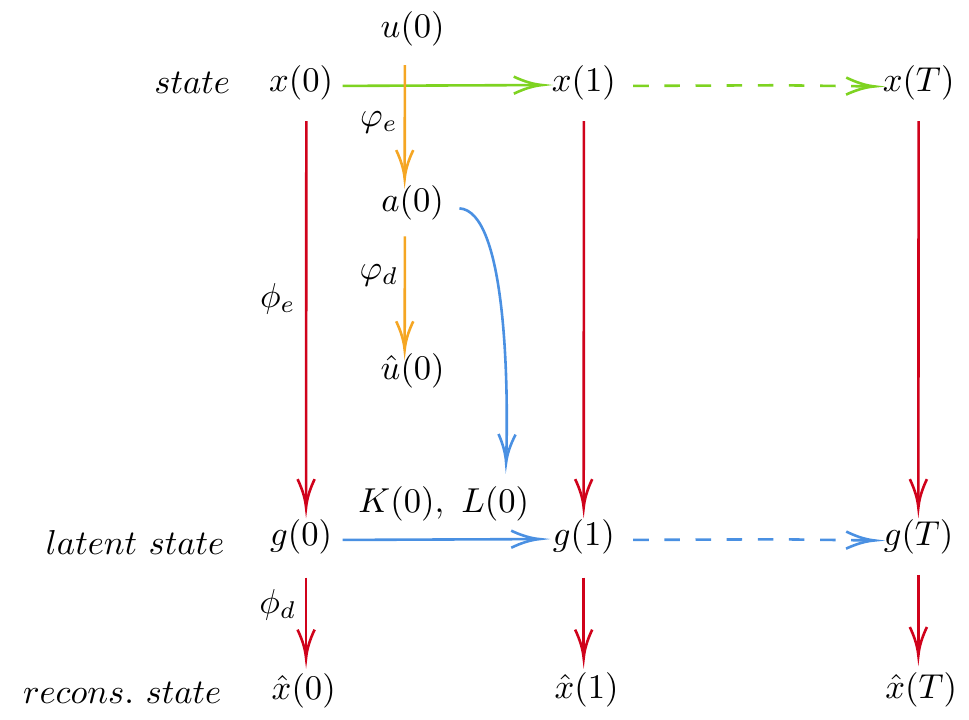}
    \caption{Diagram for prediction framework. We take a part of the diagram from time step 0 to 1 as an example. 
    The red arrows show the state autoencoder model. We use a state encoder $\phi_e$ to encode a state $x(0)$ into a latent state $g(0)$ and then use a state decoder $\phi_d$ to decode the latent state $g(0)$ into a reconstructed state $\hat{x}(0)$.
    The orange arrows show the action autoencoder model. We use an action encoder $\varphi_e$ to encode an action $u(0)$ into a latent action $a(0)$ and then use an action decoder $\varphi_d$ to decode the latent action $u(0)$ into a reconstructed action $\hat{u}(0)$.
    The green arrows show the rope dynamics. Given the state $x(0)$ and the action $u(0)$, we can have next state $x(1)$. 
    The blue arrows show the latent dynamics. Given the latent state $g(0)$, the latent action $a(0)$, state matrix $K(0)$, and control matrix $L(0)$, we can get the next latent state $g(1)$ based on locally linear latent dynamics.}
    \label{fig:prediction_framework}
\end{figure}

We consider the prediction framework in Figure~\ref{fig:prediction_framework}. It consists of the state autoencoder model, action autoencoder model, and dynamics model. 

When predicting, given a new state $x(t)$ and a new action $u(t)$, we can get latent state $g(t)$, latent action $a(t)$, state matrix $K(t)$, control matrix $L(t)$. Using locally linear latent dynamics, we can have predicted next latent state $g^{pred}(t+1)=K(t)g(t)+L(t)a(t)$. By decoding the predicted next latent state $g^{pred}(t+1)$, we will have the predicted next state $x^{pred}(t+1)$. 
%The process is the same if we want to have multi-step prediction. 
%%KS.10.31 - Do we actually need to use the encoders/decoders at every timestep for multi-step prediction? - no, we don't

\textbf{Loss Function.} 
We consider minimizing the losses for state autoencoder model, action autoencoder model, and dynamics model. 
The loss of state autoencoder model is
\begin{equation}
    \mathcal{L}_{state} = \sum_{t=0}^T ||\phi_d( \phi_e (x(t)) ) - x(t)||_2^2. 
    \label{eq:state_loss}    
\end{equation}
The state loss $\mathcal{L}_{state}$ makes sure the reconstructed state $\hat{x}$ is similar to ground truth state $x$. The loss of action autoencoder model is
\begin{equation}
    \mathcal{L}_{action} = \sum_{t=0}^T ||\varphi_d( \varphi_e (u(t)) ) - u(t)||_2^2.
    \label{eq:action_loss}      
\end{equation}
The action loss $\mathcal{L}_{action}$ makes sure the reconstructed action $\hat{u}$ is similar to ground truth action $u$. The loss of dynamics model is 
\begin{align}
    \mathcal{L}_{dyn} =& \sum_{t=0}^T || K(t)g(t) + L(t)a(t) - g(t+1) ||_2^2.
                    %   =& \sum_{t=0}^T || K(t)\phi_e (x(t)) + L(t)\varphi_e (u(t)) - \notag\\
                    %   &~~~~~~\phi_e (x(t+1)) ||_2^2.
    \label{eq:dyn_loss}                   
\end{align} 
The dynamics loss $\mathcal{L}_{dyn}$ sets a locally linear constraint on latent dynamics. The loss of prediction model is
\begin{align}
    \mathcal{L}_{pred} =& \sum_{t=0}^{T-1} || x^{pred}(t+1) - x(t+1)||_2^2.
    % =& \sum_{t=0}^{T-1} || \phi_d \Big( K(t)g(t) + L(t)a(t) \Big) - x(t+1) ||_2^2, \notag\\
    % =& \sum_{t=0}^{T-1} || \phi_d \Big( K(t)\phi_e (x(t)) + L(t)\varphi_e (u(t)) \Big) - \notag\\
    % &~~~~~~x(t+1) ||_2^2.
    \label{eq:pred_loss}
\end{align}
The prediction loss $\mathcal{L}_{pred}$ makes sure the predicted state $x^{pred}$ is similar to ground truth state $x$. The expansions of Equation~(\ref{eq:dyn_loss}) and Equation~(\ref{eq:pred_loss}) are in Appendix~\ref{appendix:formula_expansion}.
The overall training loss is
\begin{equation*}
    \mathcal{L} = \mathcal{L}_{state} + \lambda_1 * \mathcal{L}_{action} + \lambda_2 * \mathcal{L}_{dyn} + \lambda_3 * \mathcal{L}_{pred},    
\end{equation*}
where $\lambda_1$, $\lambda_2$, and $\lambda_3$ are coefficients.

\begin{figure}[!ht]
    \centering
    \includegraphics[width=0.6\columnwidth]{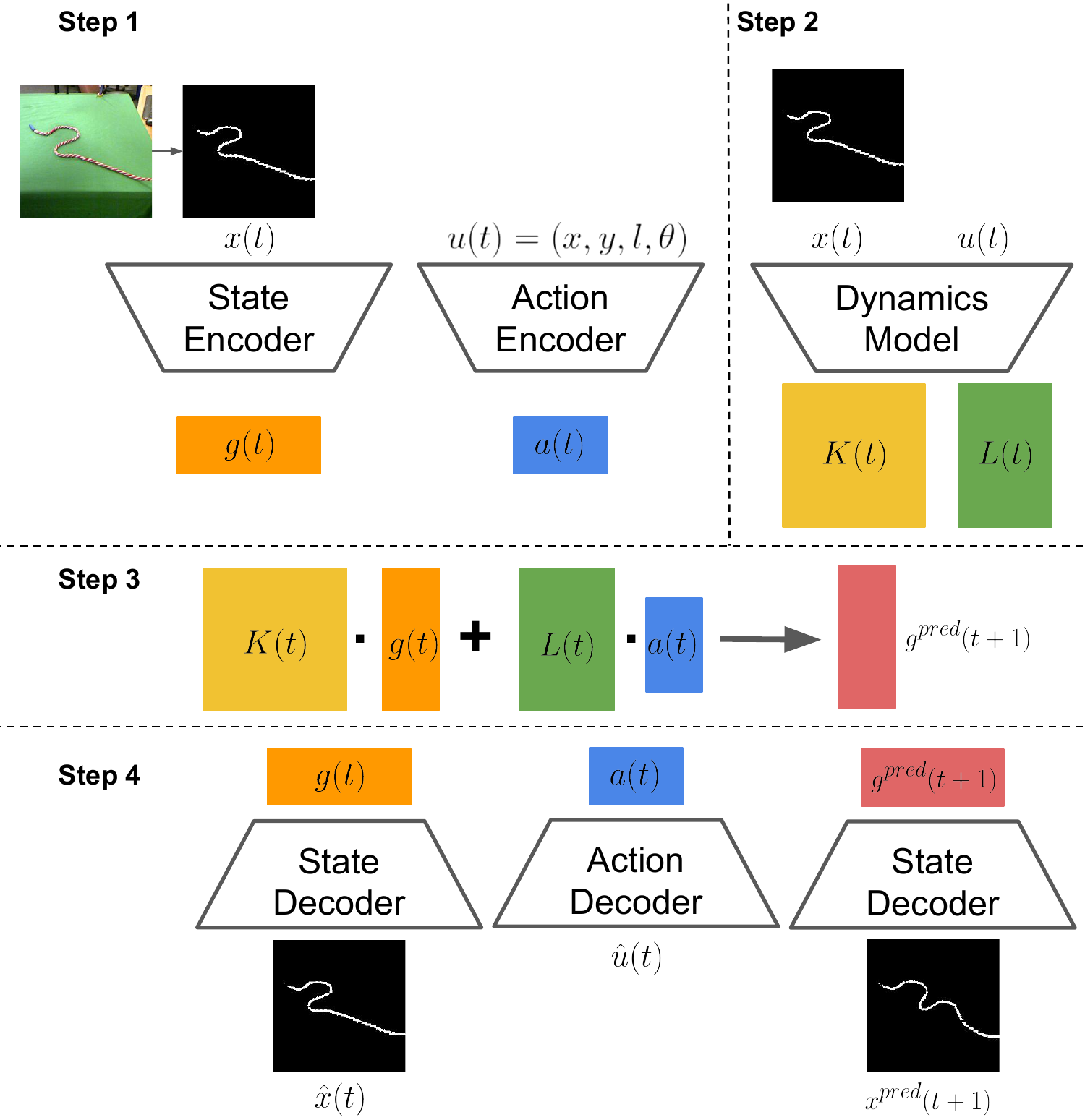}
    \caption{Prediction framework in four steps. (i) Step 1: we do color space segmentation to get a binary mask from the raw rope image. We use this mask as the input $x(t)$ for the state encoder and the output is the latent state $g(t)$ (orange vector). We also use action encoder to encode the action $u(t)$ into the latent action $a(t)$ (blur vector). (ii) Step 2: we train the dynamics model using the binary mask $x(t)$ and the action $u(t)$. The output is state matrix $K(t)$ (yellow matrix) and control matrix $L(t)$ (green matrix). (iii) Step 3: by using locally linear latent dynamics, we have predicted next latent state $g^{pred}(t+1)=K(t)g(t)+L(t)a(t)$. (iv) Step 4: we use the state decoder to decode the latent state $g(t)$ into the reconstructed state $\hat{x}(t)$, run the action decoder to decode the latent action $a(t)$ into reconstructed action $\hat{u}(t)$, and apply the same state decoder to decode the predicted next latent state $g^{pred}(t+1)$ into predicted next state $x^{pred}(t+1)$.}
    \label{fig:4steps}
\end{figure}

\textbf{Training Prediction Framework.} Figure~\ref{fig:4steps} shows 4 steps on how to train the prediction framework $\mathcal{M}$. 

\textbf{Training Autoencoders.}
In Figure~\ref{fig:4steps} Step 1, we use Equation~(\ref{eq:state_loss}) and Equation~(\ref{eq:action_loss}) to train the autoencoders to get both latent states $g$ and latent actions $a$. 
When state loss $\mathcal{L}_{state} < \epsilon_{state}$ and action loss $\mathcal{L}_{action} < \epsilon_{action}$, we save latent state $g$ and latent action $a$ in each time step. Here, $\epsilon_{state}$ and $\epsilon_{action}$ are very small values. 
%%KS.10.28 - Why?
We make sure state $x$ and action $u$ are similar to reconstructed state $\hat{x}$ and reconstructed action $\hat{u}$, respectively. 
We also do early stopping to prevent the autoencoders from overfitting to the training set.
%Appendix~\ref{sec:appendix} includes training parameters and neural network architectures for the state autoencoder and the action autoencoder. 

\textbf{Training Dynamics Model.}
The state matrix $K$ depends on state $x$ and the control matrix $L$ depends on both state $x$ and action $u$. In Figure~\ref{fig:4steps} Step 2, we train dynamics model to minimize the dynamics loss $\mathcal{L}_{dyn}$ in Equation~(\ref{eq:dyn_loss}) and prediction loss $\mathcal{L}_{pred}$ in Equation~(\ref{eq:pred_loss}).

There are two methods for model training: (i) training autoencoder models first and then training dynamics model, and (ii) training autoencoder models and dynamics model together. We discuss the model training order and our preference in Section~\ref{sec:experiment} ablation study.

To wrap up, after training the autoencoders and the dynamics model using the training set, we start to make predictions for new states and actions in the test set. The detailed architectures and the hyperparameters for the autoencoders and the dynamics model are in the Appendix~\ref{appendix:NNArchitecture} and Appendix~\ref{appendix:hyperparameter}.

\subsection{Sampling-based Model Predictive Control}
%\textbf{Sampling-based Model Predictive Control.}
% TODO: CEM backgound/application to our setup
We consider using cross-entropy method (CEM) \cite{rubinstein2013cross} to select optimal action given trained prediction framework $\mathcal{M}$. CEM is an optimization algorithm. It works by repeating two phases: (i) providing a probability distribution and sampling from this distribution and (ii) minimizing the cross-entropy between the current distribution and a target distribution for better sampling in the next iteration. Our Algorithm~\ref{alg:sample_action} is an extension of CEM. For each time step, given trained prediction framework $\mathcal{M}$, initial image $x_{init}$, and goal image $x_{goal}$, we first initialize a uniform distribution $Q_{pre}$. Then, we sample $m$ action sequences from distribution $Q_{pre}$. Each action sequence has length $h$, meaning that there are $h$ actions in this sequence. We use the prediction framework $\mathcal{M}$ to predict the future states $x^{pred}$ given initial image $x_{init}$ and each action sequence. Assuming we can get a predicted goal image $x_{goal}^{pred}$ after taking $h$ actions, we then calculate the Binary Cross Entropy (BCE) loss between the predicted goal images $x_{goal}^{pred}$ and goal image $x_{goal}$ we want. By sorting $m$ BCE losses, we choose $n~(n < m)$ action sequences that have lowest losses. We apply multivariate normal distribution $Q_{post}$ to fit $n$ samples. If the KL divergence between $Q_{pre}$ and $Q_{post}$ is not less than a specified small value, we will repeat the same process until the distribution $Q$ converges. After distribution convergence, we execute the optimal action with the lowest loss.       
\begin{algorithm}[h!]
\caption{Sampling-based optimization algorithm}
\label{alg:sample_action}
\begin{algorithmic}
\renewcommand{\algorithmicrequire}{\textbf{Input}}
\REQUIRE prediction framework $\mathcal{M}$, initial image $x_{init}$, goal image $x_{goal}$
\FOR{$t=1,...,T$}
\STATE Initialize $Q$ with uniform distribution
\WHILE{$Q$ does not converge}
\STATE Sample $m$ action sequences with length $h$ from $Q$
\STATE Use prediction framework $\mathcal{M}$ to predict goal images $x^{pred}_{goal}$ using $m$ action sequences
\STATE Calculate loss between predicted goal image $x^{pred}_{goal}$ and goal image $x_{goal}$
\STATE Select $n$ action sequences with lowest sorted losses
\STATE Fit multi-variate normal distribution $Q$ to $n$ samples
\ENDWHILE
\STATE Execute action with lowest loss
\STATE Observe new next state 
\ENDFOR
\renewcommand{\algorithmicensure}{\textbf{Output:}}
\end{algorithmic} 
\end{algorithm} 

\section{EXPERIMENTS AND RESULTS}
\label{sec:experiment}
%%KS.10.28 - General things: 1. make sure each caption can stand on its own.  It should describe the figure and tell the reader what the takeaway is.  2. Proofread.  There are a lot of places where you are missing articles like the or when the text doesn't flow smoothly.  I tried to get the ones I saw, but I probably missed some. 3. Don't use contractions.  We want to make sure that the writing is appropriately formal.

We demonstrate that our prediction framework can predict well, and our model can be used for control. We also show that training methods and locally linear latent dynamics are critical for good performance.

% Experiments show:
% Model predicts well,
% model can be used for control
% and show that xyz components of model are critical for good performance

\subsection{Experimental Setup}
% dataset, why choose it, why it's challenging, 
%
We use Rope Manipulation Dataset from paper \cite{nair2017combining}. This dataset is challenging because (i) the rope is deformable and we do not know the rope dynamics beforehand, (ii) the images are not taken straight down and there is a less-than-ninety-degree angle between the camera and the table, and (iii) the consecutive actions are not applied on the same grasping position. These challenges combine to make it difficult to predict the rope state accurately at each time step.

\textbf{State.} Different from \cite{nair2017combining} which uses the whole image as an input, we merely want to get the rope-related information in the state. Thus, we do color space segmentation and add a black mask (Figure~\ref{fig:4steps}, Step 1) to replace the noisy background. We resize the image from $240\times240$ to $50\times50$ and apply data augmentation consisting of translation, vertical flips, and horizontal flips.

\textbf{Action.} The original action $u = (x, y, l, \theta, \mathbbm{1})$ includes $x$ and $y$ positions in the image space, moving length $l$ and moving angle $\theta$ in the world space, and an indicator $\mathbbm{1}$ that decides whether this action validly moves the rope or not. Situations for the invalid actions include the robotic gripper not contacting the rope and not moving the rope according to the command.
We re-calculate the action $x$ and $y$ positions according to the change of image size and filter the invalid actions when training and testing. 

\subsection{Multi-step Prediction Results And Comparison}
% 1. comparison between GT and predicted results
% 2. other baselines
%
\textbf{One-step Prediction.}
In Figure~\ref{fig:4steps} Step 3, we use Equation~(\ref{eq:latent_dyn}) to get predicted next latent state $g_{pred}$ based on the current state matrix $K$, the current latent state $g$, the current control matrix $L$, and the current latent action $a$. In Figure~\ref{fig:4steps} Step 4, we use the same decoder in the state autoencoder and decode the predicted next latent state $g_{pred}$ into predicted next state $x_{pred}$. 

\begin{figure}[ht]
    \centering
    \includegraphics[width=0.7\columnwidth]{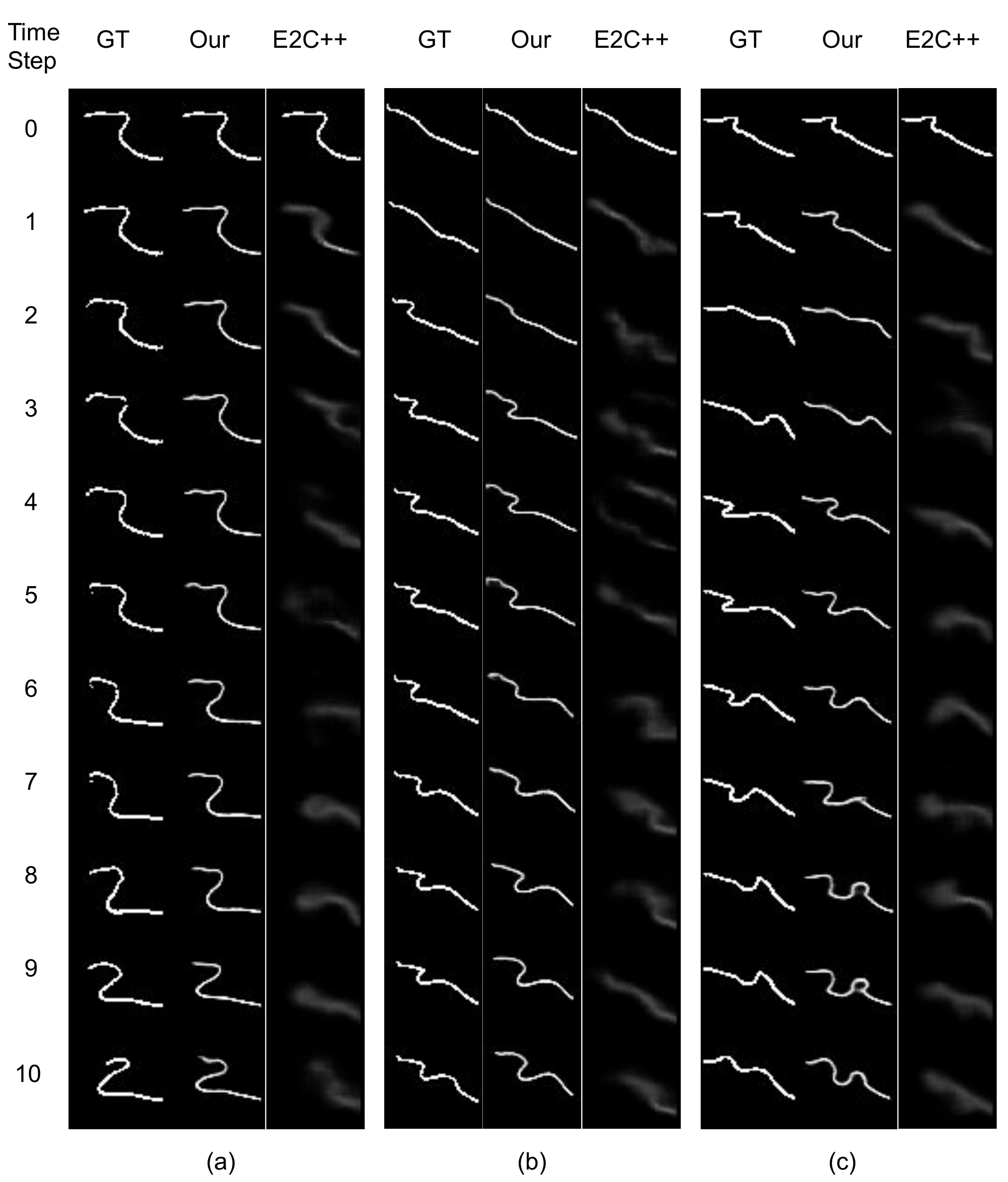}
    \caption{Comparison between Ground Truth (GT), our method (Our),  and E2C++ from time step 0 to 10. (a), (b) and (c) represent three cases starting at different initial states. In each case, it includes GT, Our, and E2C++. The first row is the initial state at time step 0. The second row (time step 1) to the last row (time step 10) are the predicted states. We can conclude that our method can predict the rope state accurately up to ten steps into the future, while the prediction results using baseline E2C++ are blurry.}
    \label{fig:10step_pred}
\end{figure}

\textbf{Multi-step Prediction.}
In order to test our framework's prediction horizon, we use the trained prediction model for multi-step prediction. Given state $x(0)$, action $u(0)$, and trained model $\mathcal{M}$, we can predict next state $x^{pred}(1)$. Then using predicted state $x^{pred}(1)$ and ground truth action $u(1)$, we predict the next state $x^{pred}(2)$. By iterating the same process, we get ten predicted states shown in Figure~\ref{fig:10step_pred}. 

\textbf{Comparison.}
We compare our results with Ground Truth and results using Embedded to Control (E2C) \cite{watter2015embed}. E2C is a method for learning and control of non-linear dynamics from images. Unlike E2C using optimal control formulation in the latent space, we use the dynamics model to get the state matrix and the control matrix for state prediction in the latent space. To make a fair comparison, we use the same neural network architecture for autoencoders and the same latent size for the state. When running the E2C on the Rope Manipulation Dataset, we find the reconstruction and prediction results are blurry, and we cannot observe a clear rope shape from the image. Thus, we add a prediction loss $\mathcal{L}_{pred}$ and a state loss $\mathcal{L}_{state}$ on the original loss of variational bound and KL divergence. We name this new approach E2C++. Figure~\ref{fig:10step_pred} (a) and (b) show that our method can predict the future states well in ten steps, and the rope shape is similar to the ground truth. For Figure~\ref{fig:10step_pred} (c), there is a clear distinction between GT and our method at time step 8 because the previous states (time step 1 to 7) gradually accumulate the state prediction error. E2C++ can only predict a blurry rope shape within two-time steps. The remaining state predictions using E2C++ do not make sense.  

\subsection{Action Planning and Control Results}
\textbf{Qualitative Results.}
% qualitative (images, pred images, overlayed actions)
%% KS.10.28 - We use the Cross Entropy Method, shown in Algorithm\ref{}, to plan rope manipulation. Add a sentence about what the objective is/any modifications you did the standard CEM, then go into talking about how you experiments are set up.
We use the sampling-based optimization method, shown in Algorithm~\ref{alg:sample_action}, to plan rope manipulation. The objective is to generate an optimal action sequence. Similar to the standard cross-entropy method (CEM), our algorithm stops repeating in the inner loop when the probability distribution $Q$ converges. However, unlike repeating two phases for the standard CEM, our algorithm needs to do more steps of sampling action sequences, predicting the goal state, calculating and sorting the loss, and fitting a new distribution. 
In the experiments, we first sample the optimal action by applying CEM. Then we use the trained prediction framework to predict the next state $x^{pred}(t+1)$ given the current state $x(t)$ and the sampled optimal action. 

\begin{figure}[h]
    \centering
    \includegraphics[width=0.7\columnwidth]{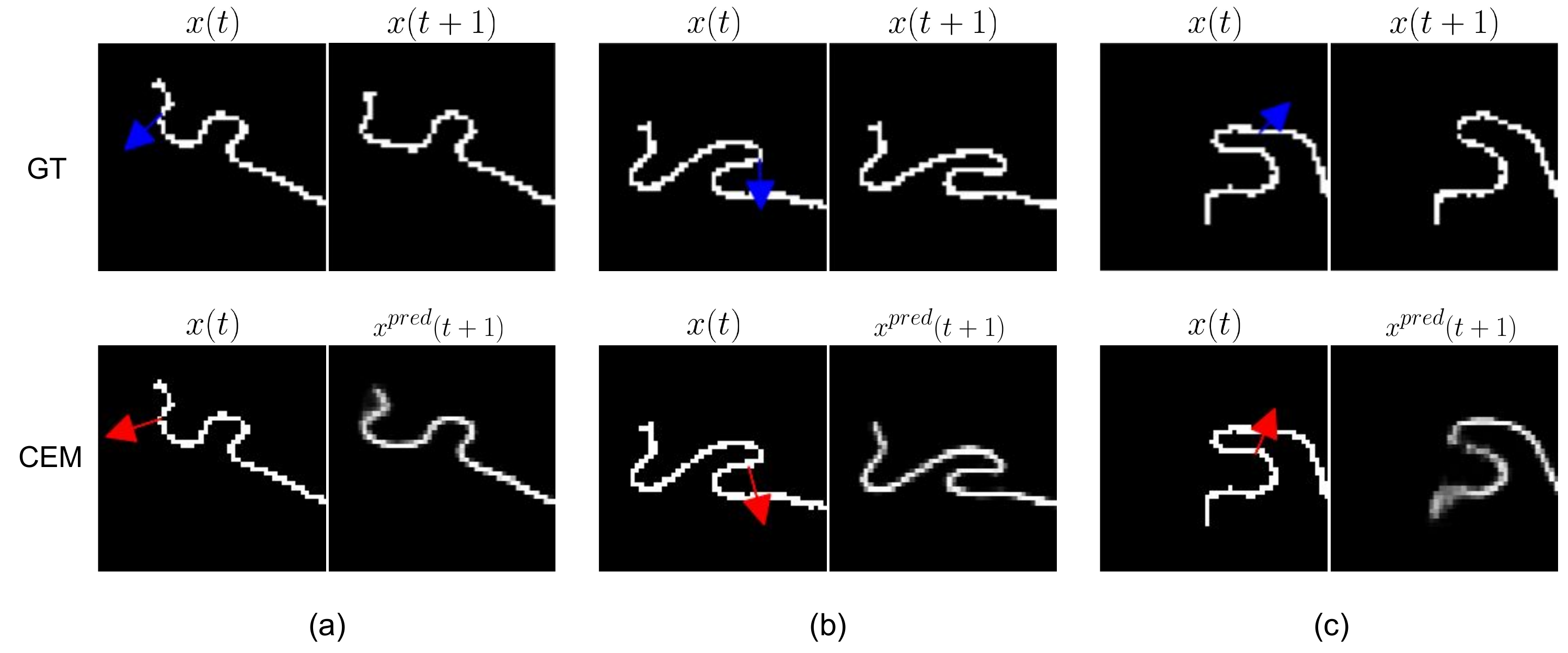}
    \caption{Comparison between Ground Truth (GT) actions and sampled actions using the Cross-Entropy Method (CEM). (a), (b) and (c) are three cases with different initial rope positions. In each case, the first row shows GT action (blue arrow) at state $x(t)$ (left) and the GT next state $x(t+1)$ (right). The second row shows sampled action (red arrow) at state $x(t)$ (left) and the predicted next state $x^{pred}(t+1)$ (right). We can conclude that the sampled actions using CEM are close to GT actions concerning the grasping positions, the moving length, and the moving angle. For case (c), sometimes there are multiple options to move the rope. Besides, our prediction model can accurately predict the next state. The predicted state $x^{pred}(t+1)$ is similar to GT state $x(t+1)$.}
    \label{fig:cem}
\end{figure}

Figure~\ref{fig:cem} shows the experimental results. We overlay sampled action on the rope and compare it with the ground truth action. By comparing the Ground Truth (GT) action with sampled action using CEM, we can find sampled action is close to GT action concerning grasping positions, moving length, and moving angle. However, sometimes there are many possible actions to move the rope from one state to the next state. 
For example, Figure~\ref{fig:cem} (c) shows the different grasping positions for sampled action, but the rope ends up with a similar predicted next state as the GT state. 

\textbf{Quantitative Results.}
We calculate the Mean Square Error (MSE) between sampled actions and ground truth actions. We also apply MSE between the predicted next state and ground truth next state. After running 100 sampled one-step actions using our method and E2C++, we get the mean and standard deviation in Table~\ref{tab:score}. By comparing Action Score, we can observe that our score is lower. It means the sampled action is closer to the ground truth action when using our prediction model. From State Score, we can find our method has a lower score. It means our method is better at predicting the next state after applying the sampled action. Detailed methods to get the quantitative results are mentioned in Appendix~\ref{appendix:state_action_score}.
\begin{table}[ht]
 \centering
 \begin{tabular}{||c c c||} 
 \hline
 Method & Action Score (MSE) & State Score (MSE)\\ [0.5ex] 
 \hline\hline
 Our & $\textbf{18.20}\pm \textbf{9.46}$ & $\textbf{0.59}\pm \textbf{0.27}$ \\ 
 \hline
 E2C++ & $29.35\pm 5.86$ & $1.12\pm 0.18$ \\ [0.51ex] 
 \hline
 \end{tabular}
 \caption{Action score and state score comparison between our method and E2C++.}
 \label{tab:score}
\end{table}
% \begin{center}
%  \begin{tabular}{||c c c||} 
%  \hline
%  Method & Action Score & State Score \\ [0.5ex] 
%  \hline\hline
%  Our & $18.20\pm 9.46$ & $0.59\pm 0.27$ \\ 
%  \hline
%  E2C++ & $29.35\pm 5.86$ & $1.12\pm 0.18$ \\ [0.51ex] 
%  \hline
% \end{tabular}
%\caption{Comparison of action scores and state scores using our method and E2C}
% \label{table:score}
% \end{center}

\subsection{Ablation Study}
\label{subsec:ablation}
\textbf{Locally Latent Dynamics v.s. Globally Latent Dynamics.}
For locally latent dynamics, state matrix $K(t)$ and control matrix $L(t)$ are different at each time step. However, $K$ and $L$ are constant matrices for globally latent dynamics. If $K$ and $L$ are constant matrices, it will set a more strict constraint in the latent space. It is difficult to train a good-performance model with globally linear latent dynamics since we need to guarantee the dynamics are linear for all latent states. In our experiments, we use locally linear latent dynamics.  

\textbf{Order for Training Models.}
We have autoencoder models for state and action and the dynamics model. The order to train models is important to get good performance. We use two methods: (i) training autoencoder models first and then training dynamics model, (ii) training autoencoder models and dynamics model together. We demonstrate that method (i) can provide better prediction and control results, as shown in Figure~\ref{fig:10step_pred}, while method (ii) reconstructs blurry rope images, and we cannot find a clear rope shape. The reason is that we can get a good latent state estimation when training autoencoder models. After having a good latent state estimation, we then train the dynamics model to get good system dynamics. For method (ii), training all models together to optimize all losses is more difficult to get a good prediction result. It is because we need to find the optimal neural network weights to get accurate latent state, latent action, state matrix, and control matrix at the same time. 

\textbf{Latent State and Latent Action Size.}
Since the dimension of state $x$ is $50\times50$, we need to shrink its size to a smaller dimension for latent state $g$. Different from the state, we use the autoencoder model to expand the action dimension from $(x,y,l,\theta)$ to a larger size. We do experiments to find how small we can get for the latent size to achieve good performance. Experiments show that the latent state size cannot be less than 50, and latent action size cannot be less than 10. When the latent state and latent action size are too small, there is no enough information about the real state and action. Besides, their sizes do not need to be too large (e.g., latent state size is 300+ and latent action size is 100+). One reason is that it increases more time to train since we have state matrix $K$ and control matrix $L$ as well. Another reason is that the larger size does not improve the prediction performance much. Thus, we choose 80 and 80 for latent state size and latent action size, respectively.

\textbf{Whether Using Action Autoencoder.}
In the prediction framework, we use the autoencoder to lift the action space to a higher dimensional space so it can work linearly with the latent state in the latent space. We also try not to use the autoencoder in the prediction framework, and we cannot achieve good prediction results due to the non-linearity of the impact of the four-element action.

%\section{CONCLUSIONS AND FUTURE WORK}
\section{CONCLUSIONS}
This paper presents a prediction framework for rope prediction. Our method first maps a non-linear space into a locally linear space using the encoders. 
Then, we get the state matrix and control matrix for linear dynamics by running the dynamics model. In the next step, we make a prediction based on the locally linear latent dynamics. 
Finally, we decode the predicted latent state into the predicted state in the non-linear space. We also propose a sampling-based optimization algorithm to select the optimal control action to move the rope from an initial state to a goal state. The experimental results demonstrate that our method can accurately predict the rope images to ten steps in the future.
The experiments also show that our sampling-based optimization algorithm can find the optimal actions to relocate the rope. By comparing sampled actions using our method and actions using the baseline, we demonstrate that our method achieves better performance.

Our approach has not demonstrated success on more complex rope configurations for linear rope prediction, such as loops or knots. Future work includes predicting more complex rope configurations, generalizing the rope prediction framework across different rope materials, and deploying our framework on a robotic manipulator.


%%%%%%%%%%%%%%%%%%%%%%%%%%%%%%%%%%%%%%%%%%%%%%%%%%%%%%%%%%%%%%%%%%%%%%%%%%%%%%%%

%%%%%%%%%%%%%%%%%%%%%%%%%%%%%%%%%%%%%%%%%%%%%%%%%%%%%%%%%%%%%%%%%%%%%%%%%%%%%%%%

% \newpage
%%%%%%%%%%%%%%%%%%%%%%%%%%%%%%%%%%%%%%%%%%%%%%%%%%%%%%%%%%%%%%%%%%%%%%%%%%%%%%%%


\section*{ACKNOWLEDGMENT}
We thank Bernadette Bucher for her contributions to this project. The research was sponsored by the Army Research Office and was accomplished under grants ARO MURI W911NF-20-1-0080, NSF CPS 2038873, ARL DCIST CRA W911NF-17-2-0181, ONR N00014-17-1-2093, and by the Honda Research Institute.

% The preferred spelling of the word ÒacknowledgmentÓ in America is without an ÒeÓ after the ÒgÓ. Avoid the stilted expression, ÒOne of us (R. B. G.) thanks . . .Ó  Instead, try ÒR. B. G. thanksÓ. Put sponsor acknowledgments in the unnumbered footnote on the first page.

%%%%%%%%%%%%%%%%%%%%%%%%%%%%%%%%%%%%%%%%%%%%%%%%%%%%%%%%%%%%%%%%%%%%%%%%%%%%%%%%

% References are important to the reader; therefore, each citation must be complete and correct. If at all possible, references should be commonly available publications.

\bibliographystyle{IEEEtran}
\bibliography{IEEEexample}

\newpage
\section*{APPENDIX}
\label{sec:appendix}
\label{sec:appendix}

\subsection{Hyperparameters}
\label{appendix:hyperparameter}
Our experimental results are based on the hyper-parameters in Table~\ref{tab:hyperparameter}.
\begin{table}[ht]
\centering
\caption{Hyperparameters for our experimental results}
\begin{tabular}{||c|c||}
\hline
    hyperparameters & value \\
    \hline\hline
    epochs (overall) & $1000$ \\
    epochs (state and action encoder-decoder) & $500$ \\
    epochs (dynamics model) & $500$ \\
    learning rate & $1\times10^{-3}$\\
    batch size & $32$ \\
    latent state size & $80$ \\
    latent action size & $80$ \\
    $\lambda_1~(\text{action coefficient in the loss function})$ & $450$ \\
    $\lambda_2~(\text{dynamics coefficient in the loss function})$ & $900$ \\
    $\lambda_3~(\text{prediction coefficient in the loss function})$ & $10$ \\   
    \hline
\end{tabular}
\label{tab:hyperparameter}
\end{table}

\subsection{Action Score And State Score}
\label{appendix:state_action_score}
We calculate the Mean Square Error (MSE) between the sampled actions and the ground truth actions. 
\begin{equation*}
\text{MSE}_\text{action} = \frac{1}{n} \sum_{i=1}^n (u_\text{sample} -u_\text{GT})^2,    
\end{equation*}
where $u_\text{sample}$ is the sampled action and $u_\text{GT}$ is the ground truth action. $n=4$ since there are four elements $(x,y,l,\theta)$ in the action $u$.

We also apply MSE between predicted next state and ground truth next state. 
\begin{equation*}
\text{MSE}_\text{state} = \frac{1}{n} \sum_{i=1}^n (x_\text{pred} - x_\text{GT})^2,    
\end{equation*}
where $x_\text{pred}$ is the predicted next state and $x_\text{GT}$ is the ground truth state. $n=2500$ since the state is a $50\times50$ image. To get the MSE for the state, we first do pixel-wise subtraction between $x_\text{sample}$ and $x_\text{GT}$. Then we square each pixel value in the image matrix. After that, we sum up all the pixel values and divide it by the image size $n$.   

After applying different sampled one-step actions on different states and running it 100 times based on our method and E2C++, we get the mean and standard deviation for the State Score and the Action Score.

\subsection{Loss Function Expansion}
\label{appendix:formula_expansion}
The expansion of the dynamic loss is
\begin{align*}
    \mathcal{L}_{dyn} =& \sum_{t=0}^T || K(t)g(t) + L(t)a(t) - g(t+1) ||_2^2, \notag\\
                      =& \sum_{t=0}^T || K(t)\phi_e (x(t)) + L(t)\varphi_e (u(t)) - \notag\\
                       &~~~~~~\phi_e (x(t+1)) ||_2^2.
    %\label{eq:dyn_loss}                   
\end{align*} 
The expansion of the prediction loss is
\begin{align*}
    \mathcal{L}_{pred} =& \sum_{t=0}^{T-1} || x^{pred}(t+1) - x(t+1)||_2^2, \notag\\
    =& \sum_{t=0}^{T-1} || \phi_d \Big( K(t)g(t) + L(t)a(t) \Big) - x(t+1) ||_2^2, \notag\\
    =& \sum_{t=0}^{T-1} || \phi_d \Big( K(t)\phi_e (x(t)) + L(t)\varphi_e (u(t)) \Big) - \notag\\
    &~~~~~~x(t+1) ||_2^2.
    %\label{eq:pred_loss}
\end{align*}

\newpage
\subsection{Neural Network Architecture}
\label{appendix:NNArchitecture}
\begin{figure}[h]
    \centering
    \includegraphics[width=\columnwidth]{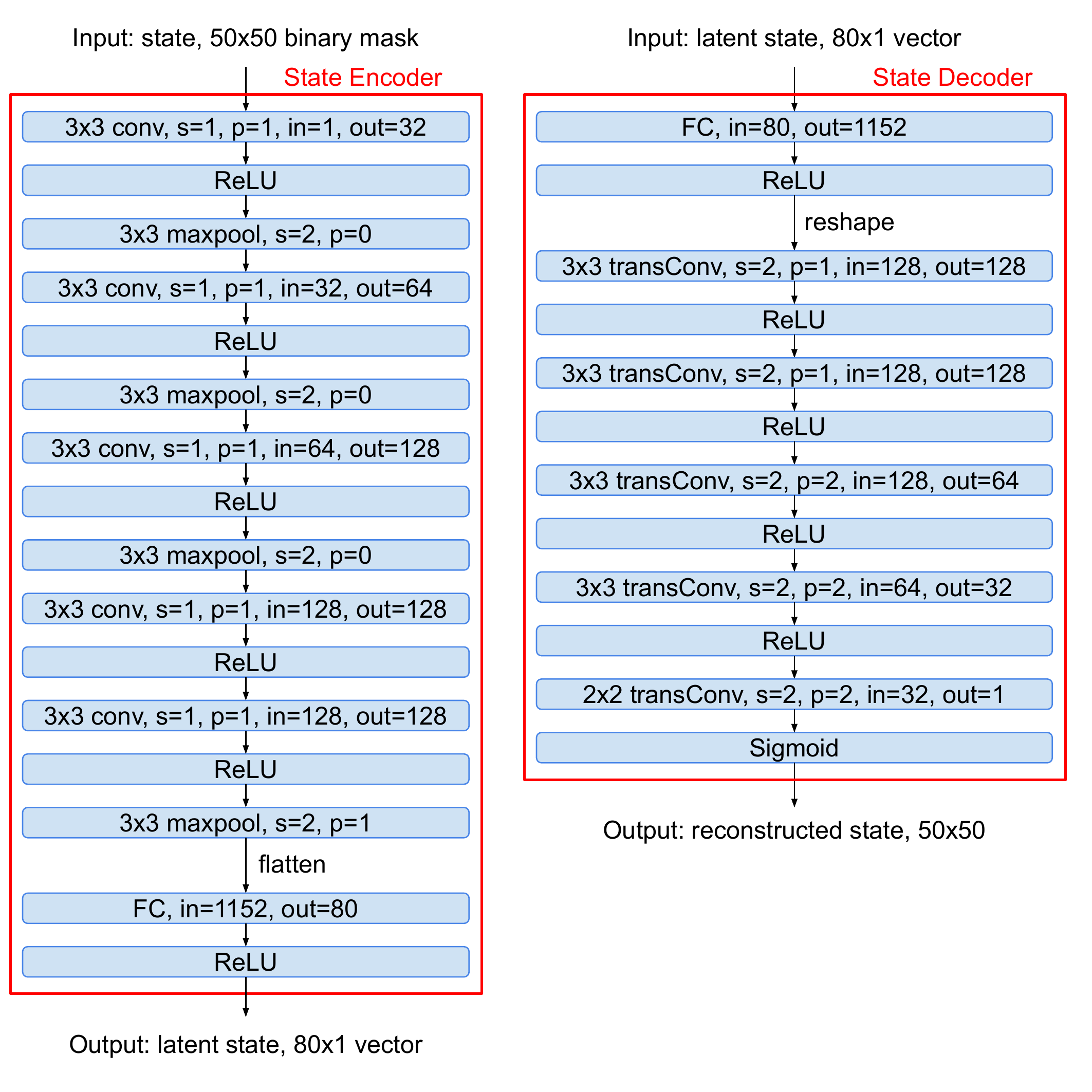}
    \caption{Neural network architectures for the state encoder and the state decoder. For the state encoder, the input state $x$ is a $50\times50$ binary mask and the output is an $80\times1$ vector for the latent state $g$. For the state decoder, the input is an $80\times1$ vector for the latent state $g$ and the output is a $50\times50$ matrix for the reconstructed state $\hat{x}$. $3\times3$ conv means the convolutional layer with the kernel size $3\times3$. The transConv means the transposed convolutional layer. In some layers, s is for the stride, p is for the padding, in means the number of input channels, out means the number of output channels, and FC means the fully connected layer.}
    \label{fig:state_encoder_decoder}
\end{figure}

\begin{figure}[h]
    \centering
    \includegraphics[width=\columnwidth]{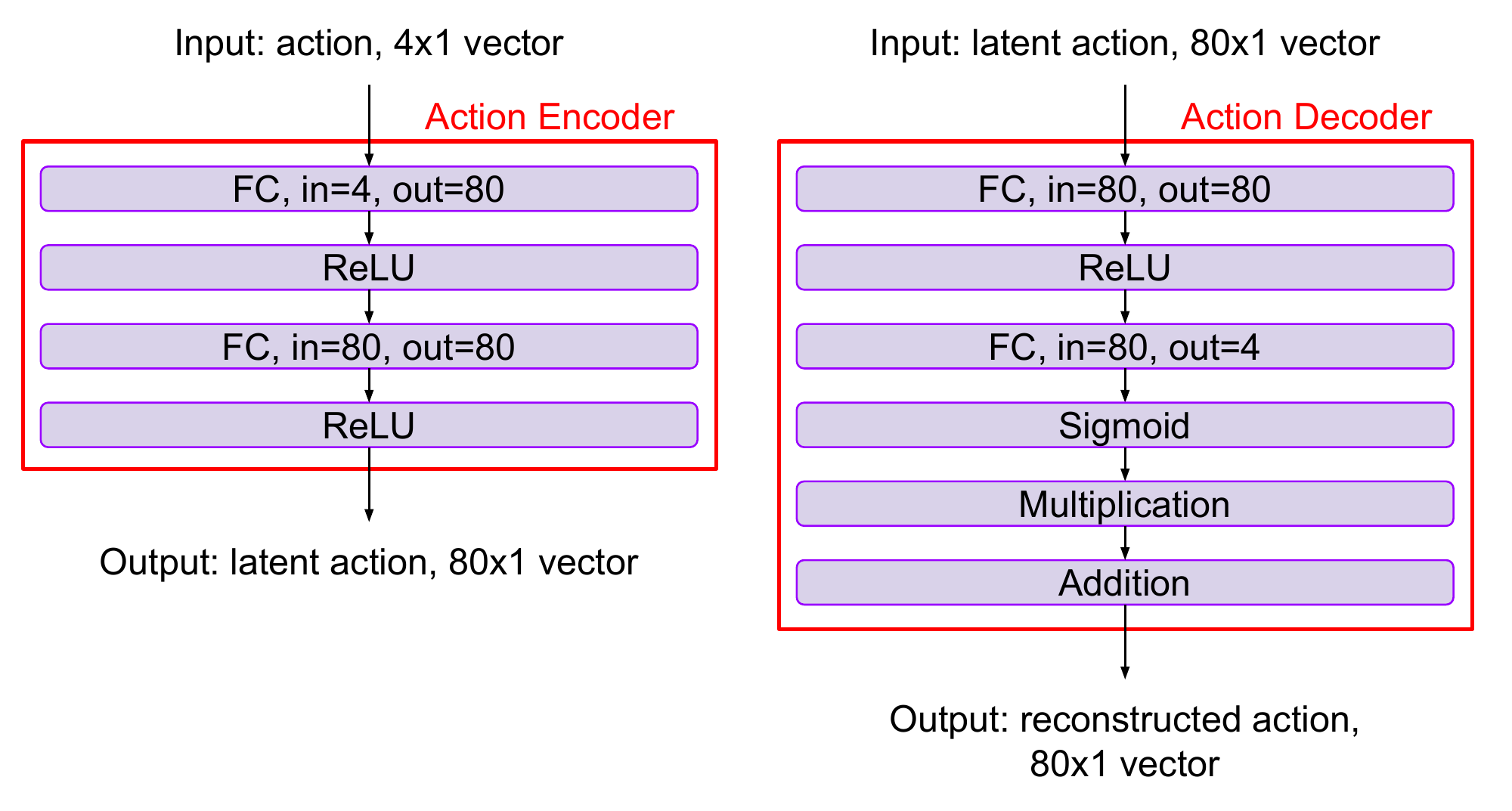}
    \caption{Neural network architectures for the action encoder and the action decoder. For the action encoder, the input action $u=(x, y, l, \theta)$ is a $4\times1$ vector, where $x$ and $y$ are the grasping positions in the image space, $l$ is the moving length, and $\theta$ is the moving angle. The output is an $80\times1$ vector for the latent action. In some layers, FC means the fully connected layer, in means the number of input features, and out means the number of output features. ReLU and Sigmoid are the activation functions. For the Multiplication layer, we do element-wise multiplication between the input and the tensor $[50,50,0.14,2\pi]$. For the Addition layer, we do element-wise addition between the input and the tensor $[0, 0, 0.01, 0]$.}
    \label{fig:action_encoder_decoder}
\end{figure}

\begin{figure}[h]
    \centering
    \includegraphics[width=\columnwidth]{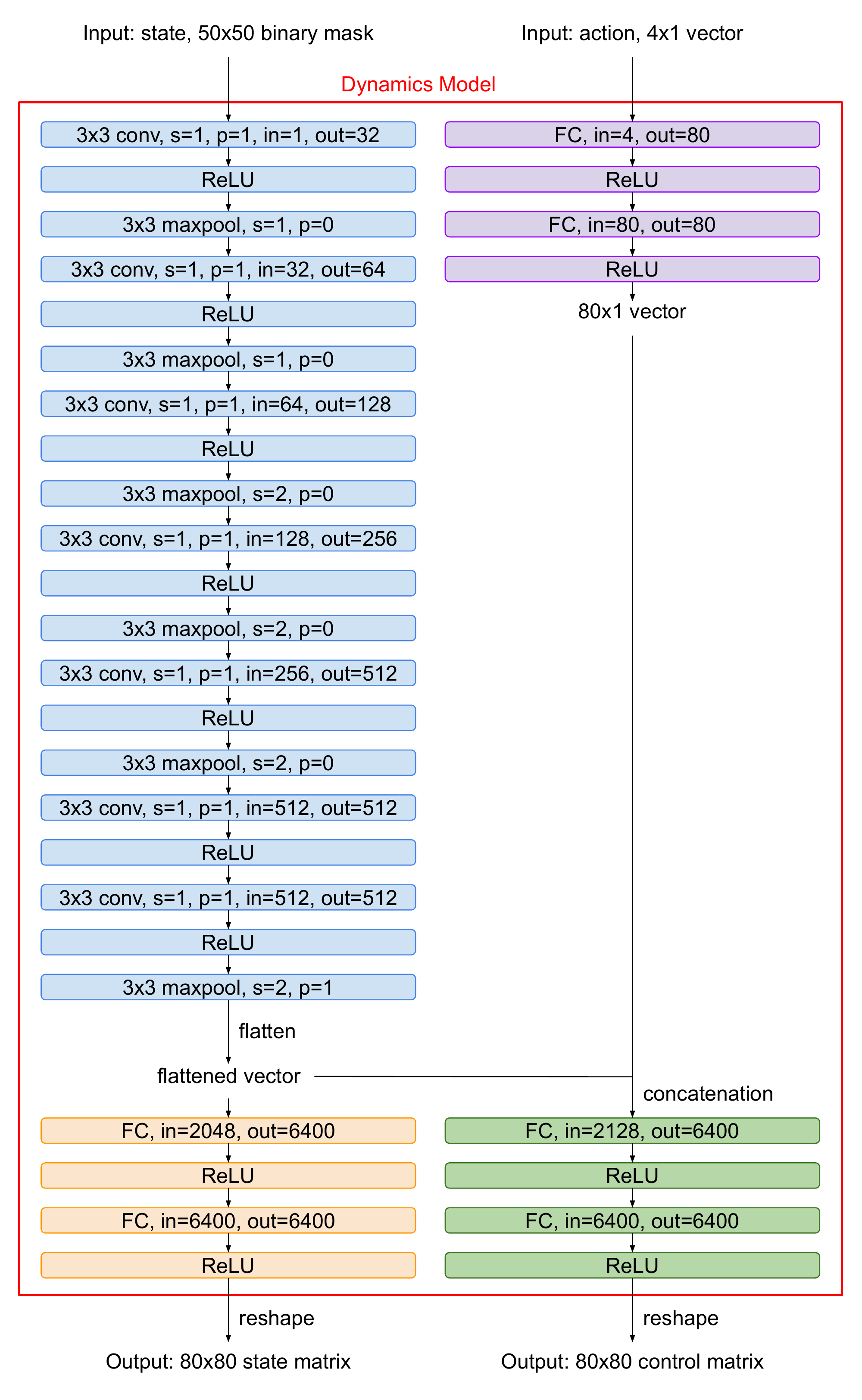}
    \caption{Neural network architecture for the dynamics model. The inputs are the $50\times50$ binary mask for the state $x$ and the $4\times1$ vector for the action $u$. In the dynamics model, there are two mappings. In the blue layers, we apply the convolutional (conv) layer followed by the activation function ReLU and max pooling for several times. After flattening the output from the blue layers, we use the flattened vector as the input for the orange layers. In the orange layers, we apply fully connected (FC) layers and activation function ReLU. After that, we reshape the vector into an $80\times80$ state matrix $K$. For the input action $u$, we first use FC and activation function ReLU to lift the action into an $80\times1$ vector in the high-dimensional space. Then we concatenate this vector with the flattened vector from the blue layers. Finally, we use the concatenated vector as an input to the green layers and reshape the output into an $80\times80$ control matrix. In the convolutional layers, s is for the stride, p is for the padding, in is for the number of input channels, and out is for the number of output channels. In the FC layers, in is for the number of input features and out is for the number of output features.}
    \label{fig:dynamics_model}
\end{figure}

Our experimental results are based on the following neural networks. Figure~\ref{fig:state_encoder_decoder} includes the input, output, and details of each layer for both the state encoder and the state decoder. Figure~\ref{fig:action_encoder_decoder} introduces the action encoder and the action decoder. In the action decoder, the Multiplication and Addition layers make sure that each element (position $x$, position $y$, moving length $l$, moving angle $\theta$) in the action are within certain range. We multiply tensor $[50, 50, 0.14, 2\pi]$ to the output from the Sigmoid layer and then add another tensor $[0, 0, 0.01, 0]$. The position $x\in[0, 50]$, the position $y\in[0, 50]$, the moving length $l\in[0.01, 0.15]$, and the moving angle $\theta\in[0, 2\pi]$. Figure~\ref{fig:dynamics_model} includes two mappings. One mapping is from the state $x$ to the state matrix $K$. The other mapping is from the state $x$ and the action $u$ to the control matrix $L$.

\newpage

\end{document}